\documentclass{article}
\usepackage{ijcai20}

\usepackage{times}
\usepackage{soul}
\usepackage{url}
\usepackage[hidelinks]{hyperref}
\usepackage[utf8]{inputenc}
\usepackage[small]{caption}
\usepackage{graphicx}
\usepackage{amsmath}
\usepackage{amsthm}
\usepackage{booktabs}
\usepackage{algorithm}
\usepackage{algorithmic}
\usepackage{amsfonts}
\usepackage{enumitem}
\usepackage{latexsym}
\usepackage{autobreak}
\usepackage{subfigure}
\usepackage{bbding}
\usepackage{pifont}
\usepackage{placeins}
\usepackage{footmisc}

\graphicspath{{./images/}}

\title{Temporal Convolutional Attention-based Network For Sequence Modeling}

\author{
Hongyan Hao$^1$\footnotemark[1]\and
Yan Wang$^2$\footnotemark[1]\and
Siqiao Xue$^3$\and
Yudi Xia$^4$\and
Furao Shen$^1$\And
Jian Zhao$^5$
\affiliations
$^1$Department of Computer Science and Technology, Nanjing University, Nanjing, China\\
$^2$School of Artificial Intelligence, Nanjing University, Nanjing, China\\
$^3$Ant Group\\
$^4$Software Institute, Nanjing University, Nanjing, China\\
$^5$School of Electronic Science and Engineering, Nanjing University, Nanjing, China\\
\emails
\{haohy, yanwang, xyd\}@smail.nju.edu.cn, \\
\{frshen, jianzhao\}@nju.edu.cn, \\
\{siqiao.xsq\}@antgroup.com
}

\begin{document}

\maketitle

\begin{abstract}
\label{abstract}
With the development of feed-forward models, the default model for sequence modeling has gradually evolved to replace recurrent networks. Many powerful feed-forward models based on convolutional networks and attention mechanisms were proposed and show more potential to handle sequence modeling tasks. We wonder that is there an architecture that can not only achieve an approximate substitution of recurrent networks but also absorb the advantages of feed-forward models. So we propose an exploratory architecture referred to \textit{Temporal Convolutional Attention-based Network (TCAN)} which combines temporal convolutional network and attention mechanism. TCAN includes two parts, one is \textit{Temporal Attention (TA)} which captures relevant features inside the sequence, the other is \textit{Enhanced Residual (ER)} which extracts the shallow layer's important information and transfers to deep layers. We improve the state-of-the-art results of bpc/perplexity to 30.28 on word-level PTB, 1.092 on character-level PTB, and 9.20 on WikiText-2.

\end{abstract}

\footnotetext[1]{Equal Contribution}

\section{Introduction}
\label{sec:intro}

With the development of deep learning, researchers have gradually concluded three frequently used structures for sequence modeling, i.e., convolutional networks, recurrent networks and attention mechanisms. 
For the tasks of sequence learning, the "default" solutions are recurrent networks in the early years, however, there are some defects that recurrent networks are hard to avoid. Besides, the feed-forward has evolved to handle tasks of sequence modeling. We aim to explore a better architecture to implement an approximate substitution of recurrent networks using feed-forward networks. So that it can not only absorb the advantages of feed-forward models but also make up for the shortcoming of recurrent networks.

Before introducing our new architecture, we first analyze the problem of sequence modeling to conclude what characteristics an effective architecture should have. As the input of task, the sequence's data point at time step $t$ is conditioned on prior one and arbitrary two data points can be relevant. Accordingly, A feasible model should characterize causality and learn the conditional correlation between data points. Besides, for high efficiency, it should train and test data in parallel. And we hope the size of this model can be as small as possible.

The researchers have done many attempts from feed-forward models to recurrent models. In the early days of deep learning, Multi-Layer Perceptron (MLP) was regarded as a complex linear function to solve the problem of prediction. An idea found in machine learning and statistical models is that for a model, parameter sharing across different parts makes it possible to extend and generalize \cite{goodfellow2016deep}. However, MLP trains separate parameters for each input feature, that is to say, it can't share parameters. \citeauthor{DBLP:journals/tsp/WaibelHHSL89}, [\citeyear{DBLP:journals/tsp/WaibelHHSL89}] proposed a time-delay neural network (TDNN), which achieves parameter sharing by applying the same convolution kernel at each time step. But the range of memory is limited by the delay factor, so it can't handle tasks with intensive data. Recurrent neural network (RNN) is different from them, it uses the same parameters to process sequential input from the beginning to the current time step. Due to that, each member of output is the function of the previous member of output, so that RNN is theoretically capable of infinite memory \cite{DBLP:journals/corr/abs-1803-01271,goodfellow2016deep} and can process variable length input. At the same time, it satisfy causality and conditional correlation for sequence modeling, This strong expressiveness makes it be "default" choice of many sequence modeling tasks. 

Although RNNs (i.e. RNN based architectures) has much strength, two shortcomings are limiting their applicability in reality. One is that in both training and evaluation, The later time steps must wait for their predecessors to complete, this inherently sequential nature precludes parallelization in training and evaluating processes \cite{DBLP:conf/nips/VaswaniSPUJGKP17}; The other one is that with the growth of sequence's length, RNNs pay more attention to nearby context and they are sensitive to the order of words within the most recent sentence but ignore word order in the long-range context, it is to say that the "infinite memory" is unnecessary \cite{DBLP:conf/acl/JurafskyHQK18}. While some variants and optimization methods achieved significant improvements \cite{DBLP:conf/iclr/MerityKS18}, but the fundamental constraints mentioned above remain.

Recently, researchers devised many alternative feed-forward models for sequence modeling, which mainly refer to temporal convolutional network-based models (TCNs) \cite{DBLP:journals/corr/abs-1803-01271} and attention mechanism-based models \cite{DBLP:conf/nips/VaswaniSPUJGKP17}. The basic TCN architecture mainly contains three modules, causal convolution, dilated convolution and residual block \cite{DBLP:journals/corr/abs-1803-01271}. Although the output member of the $t$-th step is a function of a particular number (determined by dilation factor and kernel size) of neighboring members of input before $t$. However, TCN doesn't learn distant position’s dependency inside the sequence and it doesn't extract internal correlation information of input. 
\citeauthor{DBLP:conf/iclr/BaiKK19}, [\citeyear{DBLP:conf/iclr/BaiKK19}] propose TrellisNet, characterized by weight tying across the depth and direct injection of the input into deep layers. Besides that, it combines many optimization methods to promote performance. But they make TrellisNet bigger and slower than the original TCN. 
The representative model of attention mechanism based is Transformer \cite{DBLP:conf/nips/VaswaniSPUJGKP17} which tactfully avoids recurrence by entirely relying on attention mechanism to draw global dependencies between input and output. Based on that, researchers designed some powerful models, including GPT-2 which combines generative pre-training on a diverse corpus of unlabeled text and discriminative fine-tuning on specific tasks. The GPT-2 has achieved state of the art performance on many sequence modeling tasks. While the refinements and variants of Transformers have more strong processing power for sequential data, they are too big to take into account all the characteristics of the sequence model introduced earlier.

\begin{figure*}[htp]
\centering
\subfigure[Model overall]{
\label{fig:model_overall}
\includegraphics[width=0.4\textwidth]{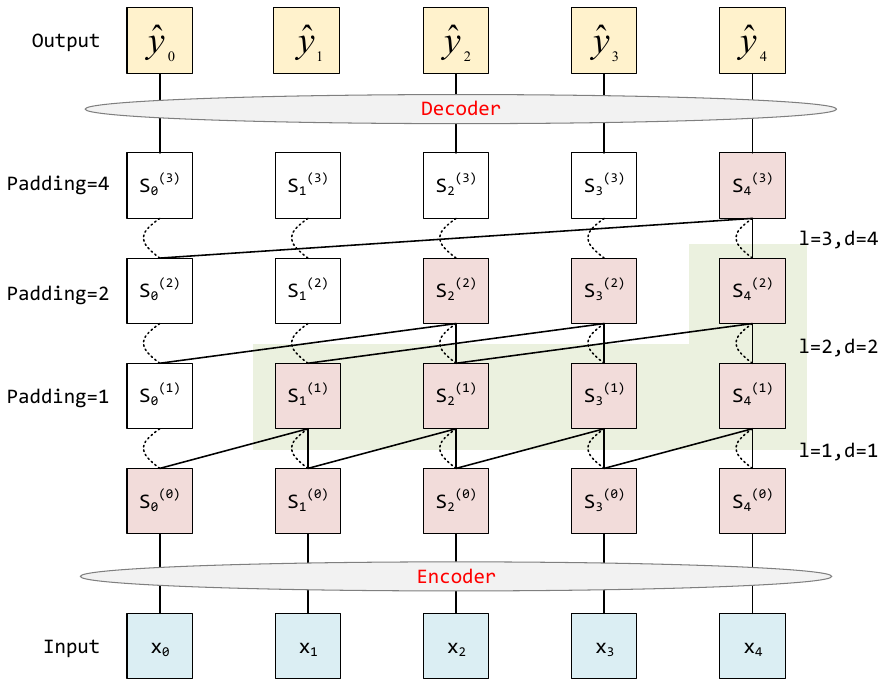}}
\hspace{.2in}
\subfigure[TCAN block]{
\label{fig:tcan_block}
\includegraphics[width=0.4\textwidth]{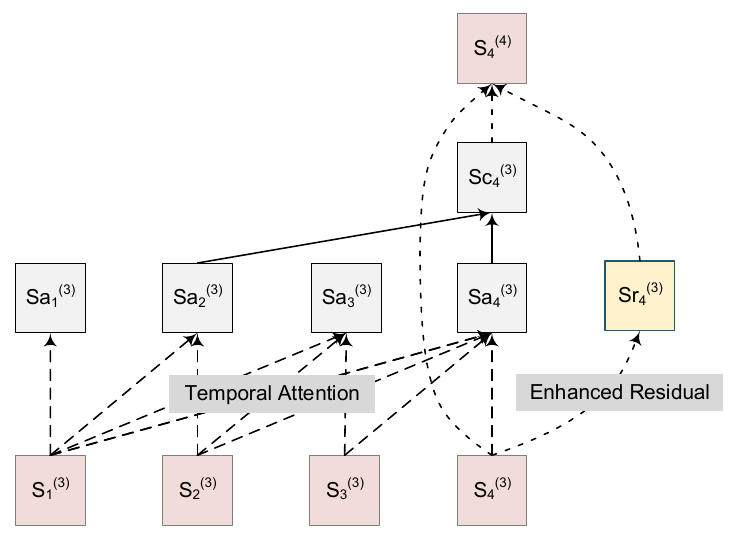}}
\caption{Architectural overall and an inter-layer transformation diagram. (a) An overall of whole architecture including Input layer, Hidden layer and Output layer, the green square indicates TCAN; (b) Inter-layer transformation of TCAN, gray squares indicate the intermediate variable of temporal attention and convolution, yellow block indicate the enhanced residual.}
\label{fig:architecture}
\end{figure*}

In this work, we propose an exploratory architecture referring to the Temporal Convolutional Attention-based Network (TCAN). On the one hand, it is inspired by TCN to utilize the dilated causal network to be an analogy to RNN's input causality. On the other hand, it combines self-attention mechanism \cite{DBLP:conf/nips/VaswaniSPUJGKP17} to extract internal correlation information and learn distant position's dependency. As illustrated in Fig~\ref{fig:architecture}, our model consists of two parts. One of them is Temporal Attention (TA) which is different from regular attention mechanism in the property of internal causality, TA's $l$-th hidden layer $t$-th time step input is a function of all previous input $x_{1:t}$, which guarantee no information leakage from future to past inside hidden layer. On the contrast, self-attention captures all time steps information in the former layer. The other part is Enhanced Residual (ER), it gets contribution weights of every time step for prediction from TA. The ER will be summed with a normal skip connection as the final residual of the current layer. In summary, we try to use a convolutional network and attention mechanism to design an approximate substitution of RNN to satisfy the conditions of solving sequence modeling.

For evaluating the impact of TCAN, we conduct experiments on the Penn Treebank (PTB) \cite{DBLP:journals/coling/MarcusSM94} and the WikiText-2 (WT2) data set \cite{DBLP:conf/iclr/MerityX0S17}. The results show that TCAN attains state-of-the-art performance by significant margins. Besides, TCAN has a smaller size than Transformer based and RNN based models. It indicates that our proposed architecture could effectively extract sequence features and could be a feasible alternative to RNN in resolving sequence modeling tasks. The code for reproducing the results is open sourced and is available at \url{https://github.com/haohy/TCAN}

\section{Methodology}
\label{sec:method}

In this section, we will introduce generic TCAN in detail. We first present the definition of the sequence modeling task in Section~\ref{ssec:seq-mod}. Then we describe the processing of data from input to output. We apply commonly used encoder-decoder structure for sequence modeling. The causal dilated network meets the requirement of sequential causality. Section~\ref{ssec:model-arch} will give an overall look at TCAN architecture. The next two sub-sections introduce the details of the two sub-modules, including Temporal Attention in Section~\ref{ssec:temp-attn} and Enhanced Residual in Section~\ref{ssec:en-res}. 

\subsection{Sequence Modeling}
\label{ssec:seq-mod}

Sequence modeling is a key problem in domains spanning audio, language modeling, music processing, time series forecasting, and many others \cite{DBLP:journals/corr/abs-1803-01271}. Suppose an input sequence as $x_{1:T} =  x_1, x_2, \cdots, x_T $ with length $T$ and the target sequence $y_{1:T} =  y_1, y_2, \cdots, y_T $ with length $T$, the task is to find a function $\mathbf{SeqMod}$ that satisfies the following relation:
$$y_{1:T} = \mathbf{SeqMod} (x_{1:T})$$ 
two constraints should be noticed: 1) $y_t$ should satisfy the causal constraint, it is a function of $x_{1:t}$, The model should prevent future information $x_{t+1:T}$ leakage; 2) the length of the input $x_{1:T}$ should be the same as the length of output. In essence, the function $\mathbf{SeqMod}$ is to find the network that minimizes some expected loss between the model's outputs and ground truth which we define it to be simply the input sequence shifted by one time step. 

Different from natural machine translation which utilizes an entire input sequence to predict the whole sentence, sequence modeling is an auto-regressive prediction which can only depend on past information. Actually, it is this constraint that makes recurrent networks become default choice for sequence modeling rather than feed-forward models.

\subsection{Model Architecture}
\label{ssec:model-arch}

From a holistic perspective, we adopt a similar structure as TCN which contains encoder and decoder like most competitive neural sequence transduction models. At the beginning of the model, the encoder maps an input sequence of symbol representations $ x_{1:T} = x_1, x_2, \cdots, x_T$ to a sequence of continuous representations $ S_{1:T}^{(0)} = \text{Encoder}(x_{1:T})$ whose T indicate the length of the sequence and 0 indicate the $0$-th layer, i.e. the first hidden layer's input. Then we apply the different kernel sizes of dilated causal convolution as a hidden layer across $L$ layers.
After the final hidden layer, the decoder generates an output sequence $\hat{y}_{1:T}=\hat{y}_1, \hat{y}_2, \cdots, \hat{y}_T$. At the most basic level, the intermediate variable at time step $t$ and level $l+1$ ($s_{1:T}^{(l+1)}$) is computed via four steps, illustrated in Figure~\ref{fig:architecture}:
\begin{enumerate}
\item 
The $s_{1:T}^{(l)}$ is passed through Temporal Attention (TA):
    $$ sa_{1:T}^{(l)} = \text{TA}(s_{1:T}^{(l)}) $$
where $sa_{1:T}^{(l)}$ indicate an intermediate variable that contains information before time steps $t$, illustrated in Figure~\ref{fig:tcan_block} and will be elaborated in Section~\ref{ssec:temp-attn}.
\item 
Given the $sa_{1:T}^{(l)}$, we apply causal convolution on it:
$$sc_{1:T}^{(l)} = \text{Conv1d}(sa_{1:T}^{(l)})$$
where $sc_{1:T}^{(l)}$ indicates the output of causal convolution. The causal block ($L_{b}$) can be stacked into many layers. For keeping the same length of each layer, we add zero padding of length ($(k-1)2^{l-1}$) on the left, white blocks in Figure~\ref{fig:model_overall}. In this way, the left relevant information of input will gradually accumulate to the right. 
\item
Before feature maps being passed through the activation function to get $s_{1:T}^{(l+1)}$, we add three components $s_{1:T}^{(l)}$, $sc_{1:T}^{(l)}$ and $sr_{1:T}^{(l)}$, where $sr_{1:T}^{(l)}$ represents Enhanced Residual (ER):
$$sr_{1:T}^{(l)} = \text{ER}(s_{1:T}^{(l)})$$
which is expressed by yellow blocks in Figure~\ref{fig:tcan_block} and will be detailed described in Section~\ref{ssec:en-res}
\item
A full TCAN network is built by stacking $L$ layers of TCAN block across depth and time, which is called dilated convolution. In addition, we use dilated convolution to enable networks to have enough receptive fields, which preserves the network computational efficiency. We set the size of dilation to increase exponentially with the depth of the network (i.e., $d = 2^l$ for layer $l$ in the network).
\end{enumerate}

\begin{figure}[h]
\centering
\includegraphics[width=0.8\linewidth]{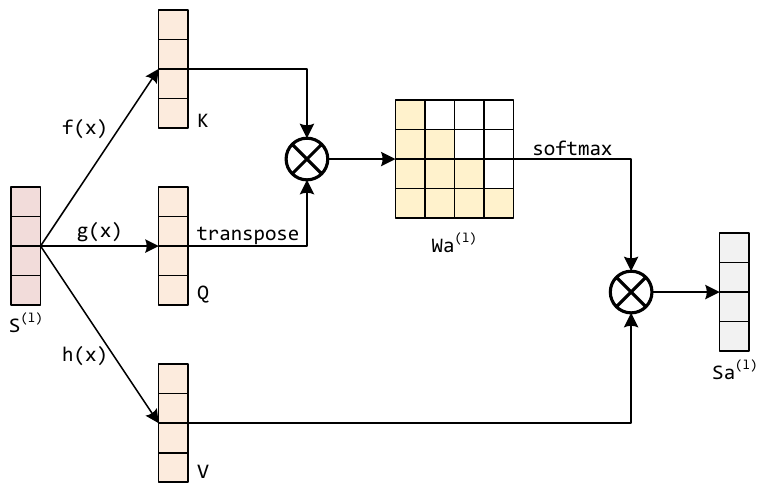}\par
\caption{Temporal Attention Block}
\label{fig:temp_attn}
\end{figure}

\subsection{Temporal Attention}
\label{ssec:temp-attn}

Temporal Attention (TA), illustrated as Figure~\ref{fig:temp_attn}, can be described as a process that integrates the influence of previous time steps into the current time step.
Our model is inspired by self-attention structure, which utilizes information of all time steps, both past and future of time step $t$. But for the sequential data, we can only handle the past information, so we refine the processing of the weight matrix to satisfy the sequential nature. 

At the first step, we use three linear transformations $f$, $g$ and $h$ to map $s_{1:T}^{(l)}$ to three different vectors, keys ($k_{1:T}^{(l)} = f(s_{1:T}^{(l)})$), query ($q_{1:T}^{(l)} = g(s_{1:T}^{(l)})$) and values ($v_{1:T}^{(l)} = h(s_{1:T}^{(l)})$) of dimension $d_k$. Then for getting the weight matrix $Wa^{(l)}$, we compute the dot products of $q_{1:T}^{(l)}$ and $k_{1:T}^{(l)}$, and divided each by $\sqrt{d_k}$. 
$$ 
W_{i,j}^{(l)} = \frac{{k_i^{(l)}}^\mathrm{T}\cdot {q_j^{(l)}}}{\sqrt{d_k}} 
$$
where $i, j = 1,2,\cdots,T$. After that, we extract the lower triangular part of $W^{(l)}$ as follows:
$$ 
Wl_{i,j}^{(l)} = 
\begin{cases} 
W_{i,j}^{(l)}, & \text{if i $\geq$ j} \\ 
0, & \text{if i $<$ j} 
\end{cases} 
$$
where $i, j = 1,2,\cdots,T$. This can shield the weight of future time steps, so as to achieve the purpose of not using future information. Finally, we apply a softmax function to normalize $Wl^{(l)}$ to get $Wa^{(l)}$, yellow blocks in Figure~\ref{fig:temp_attn}.  Given the weights, we can get the weighted output by:
$$ sa_{t}^{(l)} = \sum_{i=0}^{t} Wa_i^{(l)} \cdot s_{i}^{(l)} $$
where $t=1,2,\cdots,T$. $sa_{t}^{(l)}$ will be regarded as input of causal convolution, described in the $2$-th step in Section~\ref{ssec:model-arch}.

\begin{figure}[h]
\centering
\includegraphics[width=0.6\linewidth]{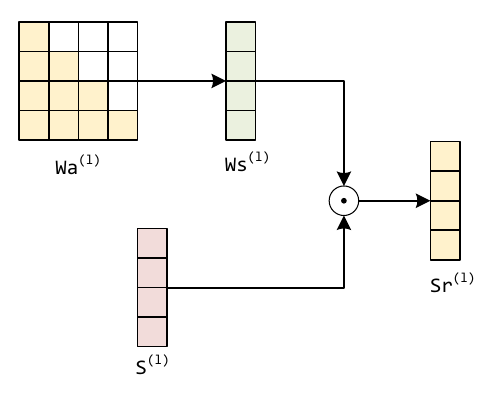}\par
\caption{Enhanced Residual Block}
\label{fig:en_res}
\end{figure}

\subsection{Enhanced Residual}
\label{ssec:en-res}
Before passing the intermediate variables through the activation function to get $s_{1:T}^{(l+1)}$, we integer three parts of information, they are identity mapping $s_{1:T}^{(l)}$, convolved vectors $sc_{1:T}^{(l)}$ and Enhanced Residual (ER) which we design to extract relatively important information and transfer them to the next layer. For a practical example of enhanced residual, in a case where you try to learn something new, if someone tells you what is the relatively important part, then you reinforce learning particular parts, it will help you grasp it faster and stronger.

Enhanced residual harness the weight matrix $Wa_{1:T}^{(l)}$ got from Temporal Attention. 
We take the sum of the weights of each row of $Wa_{1:T}^{(l)}$ to indicate the importance level of each time step, so we can get another weight vector as follows:
$$ M_t = \sum_{i=0}^{t} Wa_{i}^{(l)} $$
where $M_t$ denotes the degree of importance of time step $t$, $t=1,2,\cdots,T$. Then after the Hadamard product of $M_{t}$ and $S^{(l)}$, we get the enhanced residual ($Sr^{(l)}$).

\section{Experiments}
\label{sec:experiments}

To evaluate the effectiveness of our TCAN architecture, we conduct experiments on language modeling task (LM) task. This task is defined in Section~\ref{ssec:seq-mod}. The different datasets for LM have been described in the following section, they share the same architecture with different parameters setting. We mainly compare TCAN with representative models of three architectures listed in Section~\ref{exp:comp_methods}. Besides, we will verify the effectiveness of enhanced residual in experiments.

\subsection{Datasets and Setting}
\label{exp:datasets}

\paragraph{Penn Treebank} \cite{DBLP:journals/coling/MarcusSM94}: 
The Penn Treebank (PTB) dataset has two forms for testing the performance of models. One is word-level PTB, which contains 888K words for training, 70K for validation and 79K for testing, with a vocabulary size of 10K. Each sentence's end is marked with $<$eos$>$ and all the numbers were replaced with a $?$ symbol. The other is character-level PTB, which contains 5M characters for training, 396K for validation and 446K for testing, with an alphabet size of 50. Note that $<$eos$>$ is considered one character here. Compared to the former, character-level PTB is a medium size data set. These two forms of PTB datasets are highly studied in sequence modeling \cite{DBLP:conf/iclr/BaiKK19}\cite{DBLP:journals/corr/abs-1803-01271}\cite{DBLP:conf/iclr/KruegerMKPBKGBC17}

\paragraph{WikiText-2} \cite{DBLP:conf/iclr/MerityX0S17}:
The WikiText-2 (WT2) contains lightly pre-possessed Wikipedia articles. In contrast to PTB, it retains capitalization, punctuation and numbers. WikiText-2 features a vocabulary of over 30,000 words and contains approximately 2 million words, which is over two times larger than that of the PTB dataset.

Taking into account different characteristics of every dataset, we set different parameters for each dataset based on the same TCAN architecture. We use the gradient clip like TCN to compare the performance with TCAN based on the same optimizations. For narrative convenience, we denote the dimension of input embedding as $D_{embed}$, the number of stacking layers as $L$, the number of convolutional layers inside the TCAN as $L_b$ and the dimension of the linear transformation in Temporal Attention as $D_{attn}$. As to word-level PTB, we set the kernel size $3$. In order to make the size of our model and the comparison model as equal as possible, we set $D_{embed}=300$, $L=4$, $L_b=1$ and $D_{attn}=600$. As to character-level PTB. Because the number of characters in the whole dataset is limited and smaller than word-level PTB, we set $D_{embed}=100$. Correspondingly, because of stronger long-term dependencies between characters, we set the convolutional kernel to $7$. At the same time, the length of input sequence is larger, so we need more layers to capture the larger receptive field, we set $L=6$, $L_b=1$ and $D_{attn}=100$. As to WT2, it is a larger dataset with a twice larger size of the dictionary. so we set $D_{embed}=600$, $L=6$, $L_b=1$ and $D_{attn}=600$. we use the Adam optimizer \cite{DBLP:journals/corr/KingmaB14} with learning rate 0.0001 in all above experiments.

In order to verify the availability of enhanced residual, We consider two versions of the model, one with enhanced residual (TCAN) and one without enhanced residual (TCAN-no-res). The enhanced residual doesn't add parameters to the model, so they have equal model size.

\subsection{Compared Methods}
\label{exp:comp_methods}

\paragraph{RNN based}: Researchers proposed many variants of RNN. Among them, LSTM becomes the most used benchmark model because of solving the problem of vanishing gradients. Coupled with some regularization and optimization methods, it gets impressive results on several benchmark datasets in language modeling. AWD-LSTM \cite{DBLP:conf/iclr/MerityKS18} is a representative model, it introduces weight-dropped LSTM which uses DropConnect on hidden-to-hidden weights to be a form of recurrent regularization and NT-AvSGD which is a non-monotonically triggered (NT) variant of the averaged stochastic gradient method (AvSGD), wherein the averaging trigger is determined using a NT condition as opposed to being tuned by the user. Later, many better algorithms were produced based on the AWD-LSTM \cite{DBLP:conf/icml/WangG019,DBLP:conf/nips/GongHTQ0L18}. Besides that, We also compare TCAN with NAS models \cite{DBLP:conf/iclr/ZophL17}, which uses reinforcement learning to maximize or minimize the objective function of the generated architectures on a validation set to generate the model descriptions of neural networks.

\paragraph{CNN based}: A few notable convolutional networks have been applied to sequence modeling in recent years (e.g., the WaveNet \cite{DBLP:conf/ssw/OordDZSVGKSK16} and PixelCNN \cite{DBLP:conf/nips/OordKEKVG16} architectures). Among some derivative models, the best one is TrellisNet  \cite{DBLP:conf/iclr/BaiKK19}. It combines weight tying across depth and time and input insertion, besides that, it equips many techniques, including dropout, weight normalization and auxiliary loss. Putting these techniques together, it achieves a state-of-the-art performance among various derivative models of TCN.

\paragraph{Attention mechanism}: In recent years, the most frequently used and effective models in industry are based on Transformer \cite{DBLP:conf/nips/VaswaniSPUJGKP17} structure, which is a derivative framework of attention mechanism in essence. GPT-2 \cite{radford2019language} is considered the best model at present, and it achieves state-of-the-art results on many sequence modeling benchmark tasks. GPT-2's goal is to design a multitask learner, and it utilizes a combination of pre-training and supervised finetuning to achieve more flexible forms of transfer. Therefore it has 1542M parameters, much bigger than other comparative models.

\begin{table}
\centering
\begin{tabular}{ccc}
\hline 
\multicolumn{3}{c}{Word-level Penn Treebank (PTB)}  \\
\hline
Models & Size & $\text{ppl}^l$ \\ 
\hline
Generic TCN \cite{DBLP:journals/corr/abs-1803-01271} & 13M & 88.68 \\  
NAS Cell \cite{DBLP:conf/iclr/ZophL17} &  54M &  62.4 \\
AWD-LSTM \cite{DBLP:conf/iclr/MerityKS18} & 24M & 58.8 \\  
TrellisNet \cite{DBLP:conf/iclr/BaiKK19} & 33M & 56.80 \\ 
TrellisNet-MoS \cite{DBLP:conf/iclr/BaiKK19} & 34M & 54.19 \\ 
GPT-2 \cite{radford2019language} & 1542M & 35.76  \\ 

\hline 
TCAN-no-res & 13M & $\textbf{32.19}$  \\
TCAN & 13M & $\textbf{30.28}$ \\
\hline
\end{tabular}
\caption{Test perplexities (ppl) on word-level language modeling with the PTB corpus. $l$ means lower is better.}
\label{table:word-ptb}
\end{table}

\begin{table}
\centering
\begin{tabular}{ccc}
\hline 
\multicolumn{3}{c}{WikiText-2 (WT2)}  \\
\hline
Models & Size & $\text{ppl}^l$ \\ 
\hline
Generic TCN \cite{DBLP:journals/corr/abs-1803-01271} & 28.6M & 138.5 \\  
AWD-LSTM \cite{DBLP:conf/iclr/MerityKS18}$^{\dagger}$ & 33M & 44.3 \\  
AWD-LSTM-MoS \cite{DBLP:conf/iclr/YangDSC18}$^{\dagger}$ & 35M & 40.68 \\
GPT-2 \cite{radford2019language} & 1542M & 18.34 \\ 
\hline 
TCAN-no-res & 33M & $\textbf{10.92}$  \\
TCAN & 33M & $\textbf{9.20}$ \\
\hline
\end{tabular}
\caption{Test perplexities (ppl) on word-level language modeling with the WikiText-2 corpus. $\dagger$ indicates using dynamic evaluation.}
\label{table:wikitext-2}
\end{table}

\begin{table}
\centering
\begin{tabular}{ccc}
\hline 
\multicolumn{3}{c}{Character-level Penn Treebank (PTB)}  \\
\hline
Models & Size & $\text{ppl}^l$ \\ 
\hline
Generic TCN \cite{DBLP:journals/corr/abs-1803-01271} & 3.0M &  1.31 \\  
IndRNN \cite{DBLP:conf/cvpr/0005LCZG18} & 12.0M & 1.23 \\
NAS Cell \cite{DBLP:conf/iclr/ZophL17} & 16.3M & 1.214 \\
AWD-LSTM \cite{DBLP:conf/iclr/MerityKS18} & 13.8M & 1.175 \\  
TrellisNet-MoS \cite{DBLP:conf/iclr/BaiKK19} & 13.4M & 1.158 \\ 

\hline 
TCAN-no-res & 4.3M & $\textbf{1.104}$  \\
TCAN & 4.3M & $\textbf{1.092}$ \\
\hline
\end{tabular}
\caption{Test bits-per-character (bpc) on character-level language modeling with the PTB corpus.}
\label{table:char-ptb}
\end{table}

\subsection{Results and Analysis}
\label{exp:results_analysis}

We evaluate TCAN on word-level and character-level language modeling on Penn Treebank (PTB) and WikiText-2 datasets as mentioned in Section~\ref{exp:datasets}. The prior state of the art on these datasets are set most by GPT-2. For the word-level datasets, we use PTB and wikitext-2. The former one is a small but frequently used dataset. We also conduct some further ablation studies on it. The latter one is larger, so it reduces the risk of overfitting to a certain extent. 

As illustrated in Table~\ref{table:word-ptb} and Table~\ref{table:wikitext-2}, compared with TCAN, AWD-LSTM and TrellisNet have more parameters but poorer performance (higher perplexity). GPT-2 has better performance than others, but its model size is so big. It causes that it needs so much natural language corpus and computational resources to get the pre-trained model. Actually, the largest part of parameters is the embedding module whose size is proportional to dictionary size. So that the model size of wikitext-2 is bigger than that of word-level PTB. On PTB dataset, TCAN uses fewer parameters to achieve state-of-the-art performance. As to wikitext-2 dataset, TCAN uses the same amount of parameters as AWD-LSTM and sets a new state of the art as well. For the  character-level dataset, illustrated in Table~\ref{table:char-ptb}, we also set a new state of the art. Due to the GPT-2 was trained for word-level models, it doesn't work for this dataset. From the comparison between AWD-LSTM with TrellisNet and TCN, we found that for this task of sequence modeling, the feed-forward model can not only outperforms RNN based model, but also has a similar or even smaller size. Furthermore, TCAN has a simpler structure and better performance than TrellisNet, which suggests that the combination of TCN structure and attention mechanism is a feasible exploration.

According to the comparison between TCAN-no-res and TCAN in the chart, we found that the enhanced residual can improve the performance in some extent. As mentioned in Section~\ref{ssec:en-res}, we think that enhanced residual select valuable features and strengthen the memory of important information. Note that the values shown on Table~\ref{table:word-ptb}, \ref{table:wikitext-2} and \ref{table:char-ptb} are not the best performance results, but set for comparison.

\subsection{Ablation Experiments}
\label{exp:ablation}

To better explain the relative importance of the several conclusions we proposed, we designed a ablation experiment on the word-level PTB dataset, which is to prove that temporal attention (TA) layer is more effective than a convolutional layer. 

\begin{table}
\centering
\begin{tabular}{cccccc}
\hline 
ER & TA & $L_{b}$ & $L$ & Size & ppl$^l$ \\ 
\hline \hline 
\ding{55} & \ding{51} & 1 & 4 & 13.2M & 36.85 \\
\ding{55} & \ding{55} & 2 & 4 & 14.7M & 151.98 \\

\hline
\end{tabular}
\caption{Comparative tests on temporal attention layer and convolutional layer.}
\label{table:conv_attn}
\end{table}

When we compare the efficiency of temporal attention (TA), we discard all TA layers and replace them with convolutional layers. To guarantee the fairness of models, we adjust hyperparameters to make the convolutional model have more parameters than TCAN. The initial TCAN has 1 temporal attention block in each of the 4 layers. For comparison, we use convolutional layer to replace TA. A TA layer has similar number of parameters with a convolutional layer. Note that TCAN didn't use the enhanced residual module, which is in order to not interfere with comparative tests. The results are listed in Table~\ref{table:conv_attn}. Note that two models are optimized by Adam and the learning rate is 0.0001. The perplexity of two models shows that the temporal attention layer is more effective than the convolutional layer.

\section{Conclusion}
\label{conclusion}

In this work, we propose an exploratory architecture TCAN for sequence modeling. The sub-module temporal attention can integer internal correlative features under the condition of satisfying sequential characteristics. The other enhanced residual utilize the weight of temporal attention to emphasize the importance of certain time step, it can improve the performance without adding parameters. Our model outperforms prior state-of-the-art models on WikiText-2, word- and character-level PTB datasets by a significant margin.

\bibliographystyle{named}
\bibliography{ijcai20}

\end{document}